\documentclass{article}

\usepackage{microtype}
\usepackage{graphicx}
\usepackage{subfigure}
\usepackage{booktabs} 
\usepackage{amssymb}
\usepackage{amsmath}

\usepackage{hyperref}

\usepackage[accepted]{icml2021}

\icmltitlerunning{Multiclass Unsupervised Anomaly Detection on Masked Objects}

\begin{document}

\twocolumn[
\icmltitle{ODDObjects: A Framework for Multiclass Unsupervised Anomaly Detection on Masked Objects}



\icmlsetsymbol{equal}{*}

\begin{icmlauthorlist}
\icmlauthor{Ricky Ma}{ubc}
\end{icmlauthorlist}

\icmlaffiliation{ubc}{Department of Computer Science, University of British Columbia, Vancouver, BC, Canada}

\icmlcorrespondingauthor{Ricky Ma}{ricky.ma@alumni.ubc.ca}

\icmlkeywords{Machine Learning, ICML}

\vskip 0.3in
]

\begin{abstract}
This paper presents a novel framework for unsupervised anomaly detection on masked objects called ODDObjects, which stands for \textit{Out-of-Distribution Detection on Objects}. ODDObjects is designed to detect anomalies of various categories using unsupervised autoencoders trained on COCO-style datasets. The method utilizes autoencoder-based image reconstruction, where high reconstruction error indicates the possibility of an anomaly. The framework extends previous work on anomaly detection with autoencoders, comparing state-of-the-art models trained on object recognition datasets. Various model architectures were compared, and experimental results show that memory-augmented deep convolutional autoencoders perform the best at detecting out-of-distribution objects.
\end{abstract}

\section{Introduction}
An anomaly is a data-point that is significantly different from the rest of the data. Anomaly detection is an important task that has various applications in fraud detection, video surveillance, novelty detection, manufacturing systems, and more \cite{chandola2009}. In manufacturing systems, anomaly detection is becoming increasingly important to enhance reliability and resiliency in the modern Industry 4.0 framework \cite{pittino2020}. Advanced production monitoring systems aim to detect anomalies in real-time for quality, control, and maintenance purposes with unsupervised anomaly detection, putting humans out-of-the-loop \cite{stojanovic2016, SCIME2018114, SUSTO20172018}. It has also been shown that removing anomalous data from training data used for supervised learning often results in a statistically significant increase in model accuracy \cite{smith2011}.

Unsupervised anomaly detection \cite{zimek2012, zong2018deep} is a subset of anomaly detection that tries to learn anomalies from unlabeled data. Unsupervised models automatically identify samples that lie outside of the normal distribution of the data, which is difficult due to the lack of human supervision. The problem is even more difficult with high-dimensional data, such as images, as modeling high-dimensional data has been shown to be especially challenging \cite{zimek2012}.

Autoencoders, along with its convolutional and variational variants, are powerful unsupervised models used to learn high-dimensional data \cite{bengio2007, kingma2014autoencoding}. An autoencoder consists of an encoder and decoder. The encoder maps the high-dimensional input data into a lower-dimensional latent representation, and essentially acts as a bottleneck that forces the model to learn the most important features in representing the data. The decoder reconstructs the high-dimensional data from its latent representation. To use autoencoders for anomaly detection, the original input is compared with the output from the decoder to compute the reconstruction error, used as an indicator for anomalies. It is assumed that a higher reconstruction error corresponds to abnormal data, since they are further away from the training data, while reconstruction error remains low for normal data \cite{zong2018deep}.

\section{Related Work}
\subsection{Anomaly detection}
Anomaly detection methods can be generally classified into probability, proximity, and information theory based models \cite{outlieranalysis}.

Probabilistic anomaly detection assumes that data comes from a certain probability distribution, and the parameters of the model are learned. Gaussian mixture models \cite{reynolds2009gaussian} are parametric probabilistic models that are commonly used for anomaly detection, where normal data has a high probability of occurring and abnormal data has a low probability of occurring. Probabilistic models can be applied to almost any data-type, as long as there exists a good choice of distribution to model the data. This is also a drawback, as it may not always be appropriate to fit a model to a particular distribution.

Proximity based anomaly detection assumes that abnormal data is separated from normal data, determined by a certain metric. These models include distance, clustering, and density-based techniques. In distance-based anomaly detection, anomalous datapoints lie far away from the normal data based on a distance measure. For example, the K-nearest neighbours algorithm uses Euclidean distance to find the datapoints with the largest KNN distances \cite{knn}. Clustering-based techniques, like K-means, detect anomalies by finding datapoints furthest away from the cluster centroids \cite{munz2007traffic}. Density-based approaches consider anomalies as datapoints lying in sparse regions of the data. In DBSCAN \cite{dbscan}, all datapoints that are not clustered by the algorithm may be considered an outlier, or an anomaly.

Information theoretic anomaly detection uses data summarization techniques such as probabilistic model parameters, clusters, or lower-dimensional representations to summarize the data. Datapoints that deviate from the summary by a certain threshold are considered anomalies. Two such models are PCA \cite{rousseeuw2011robust} and autoencoders \cite{autoencoder}; both can be used to compress and reconstruct the data. The reconstruction error is the difference between the reconstructed and original data, which is used as a metric for anomaly detection. In this case, the anomalies are simply datapoints with high reconstruction errors.

\begin{figure*}[ht]
\vskip 0.2in
\begin{center}
\centerline{\includegraphics[width=\textwidth]{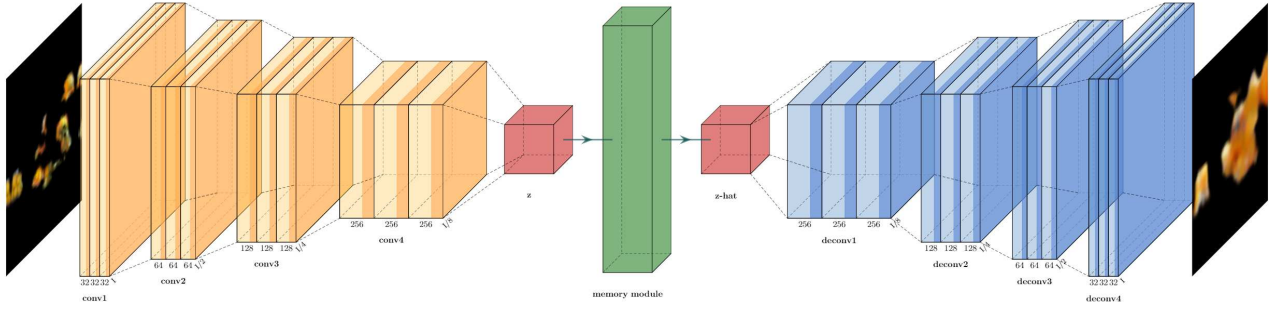}}
\caption{Memory-augmented deep convolutional autoencoder.}
\label{memcae}
\end{center}
\vskip -0.2in
\end{figure*}

\subsection{Autoencoder}
Autoencoders \cite{bengio2007} are a type of neural network with two components: an encoder and a decoder. A neural network with two hidden layers can be represented by equation (1) and (2), where $x$, $\hat{x}$, $z$, $W$, $b$, and $\sigma$ represent the input, reconstructed output, latent representation, weight, bias, and nonlinear transform function, respectively:
\begin{equation}
    z = \sigma(W_{xz}x + b_{xz})
\end{equation}
\begin{equation}
    \hat{x} = \sigma(W_{zx}z + b_{zx})
\end{equation}
The encoder, represented by equation 1, takes in the multi-dimensional input vector $x$ and compresses it to a lower-dimensional latent representation $z$. The decoder, represented by equation 2, takes the latent representation $z$ and maps it back to the original input space with the same transformation as the encoder, resulting in a reconstruction $\hat{x}$. The difference between the original input and the reconstruction is given by the reconstruction error:
\begin{equation}
    ||x-\hat{x}||
\end{equation}
By training an autoencoder, the model learns to minimize the reconstruction error. The training algorithm is shown in \textbf{Algorithm 1}, where $f_\phi$ and $g_\theta$ are multilayered neural networks for the autoencoder, parameterized by $\phi$ and $\theta$.

\begin{algorithm}[tb]
   \caption{Standard autoencoder training algorithm}
   \label{alg:autoencoder_training}
\begin{algorithmic}
   \STATE {\bfseries Input:} Dataset $\{X^i\}_{i=1}^N$
   \STATE {\bfseries Output:} Trained AE model $g_\theta$, $f_\phi$
   \STATE Initialize $\theta$, $\phi$
   \REPEAT
   \STATE $\textit{error} = \sum_i^n ||x^i - g_\theta(f_\phi(x^i))||$
   \STATE Update $\theta$ and $\phi$ via gradients of \textit{error}
   \UNTIL{convergence of $\theta$ and $\phi$}
\end{algorithmic}
\end{algorithm}

\subsection{Variational autoencoder}
In variational autoencoders \cite{kingma2014autoencoding}, the latent representation $z$ is interpreted as a latent variable in a probabilistic generative model. The decoder is instead defined probabilistically by the likelihood function $p_\theta(x|z)$, the probability of sample $x$ given the latent representation $z$. Hence, by defining a prior distribution $p(z)$ over the latent variables, the encoder is given by $p_\theta (z|x) \propto p(z)p_\theta(x|z)$. The standard normal distribution is commonly used as the prior $p(z)$, such that $z \sim q_\phi(z|x)$ is reparameterized by a deterministic transformation, resulting in a sampled $\tilde{z}$ that follows the distribution of $q_\phi(z|x)$ \cite{an2015variational}.

During training, the parameters $\theta$ and $\phi$ can be learned simultaneously via the variational Bayes approach by maximizing the evidence lower bound (ELBO):
\begin{equation}
    \mathcal{L}(\phi,\theta;x) = \mathbb{E}_{q_\phi(z|x)} [\log p_\theta(x,z) - \log q_\phi(z|x)]
\end{equation}
where $q_\phi(z|x)$ is the posterior approximation of $p_\theta (z|x)$ and $\mathcal{L}(\phi,\theta;x) \leq \log p_\theta(x)$. Training a variational autoencoder consists of sampling a value of $z$ from $q_\phi(z|x)$ to compute the ELBO, and using the ELBO to update $\theta$ and $\phi$ via gradient descent.

\subsection{Memory-augmented autoencoders}
A memory-augmented autoencoder \cite{gong2019memorizing} consists of an encoder and decoder, as well as a memory module. The encoder can be seen as a query generator, while the decoder reconstructs the inputs from retrieved memories. The memory module is made up of a memory matrix $M \in \mathbb{R}^{N \times C}$ and an attention-based addressing vector $w \in \mathbb{R}^{1 \times N}$, where $C$ is equal to the dimension of $z$. The memory network obtains $\hat{z}$ via softmax and cosine similarity operations, shown in (5) and (6).
\begin{equation}
    \hat{z} = wM = \sum_{i=1}^N w_i m_i
\end{equation}
\begin{equation}
    w_i = \frac{\exp (\text{sim}(z,m_i))}{\sum_{j=1}^N \exp (\text{sim}(z,m_j))}
\end{equation}
During training, both the $l_2$-norm based the reconstruction error (7) and the entropy of $\hat{w}^i$ (8) are minimized:
\begin{equation}
    ||x^i - \hat{x}^i||_2^2
\end{equation}
\begin{equation}
    E(\hat{w}^i) = \sum_{i=1}^N -\hat{w}_i \log(\hat{w}_i)
\end{equation}
where
\begin{equation}
    \hat{w}^i = \frac{\max (w_i - \lambda, 0) \cdot w_i}{|w_i - \lambda| + \epsilon}
\end{equation}
, $\lambda$ is the shrinkage threshold, and $\epsilon$ is a very small scalar. Hence, with a hyperparameter $\alpha$, the combined training objective for the memory-augmented autoencoder is given by equation (8).
\begin{equation}
    L(\theta_e, \theta_d, M) = \frac{1}{N} \sum_{i=1}^N (||x^i - \hat{x}^i||_2^2 + \alpha E(\hat{w}^i))
\end{equation}

Using the encoded representation as a query, the memory module retrieves the most relevant items in memory via the attention-based addressing operator. While the encoder and decoder are still optimized to minimize reconstruction error, the added memory module is continuously updated through optimization via backpropagation and gradient descent to record the latent representations of input data. The memory module, therefore, aids the encoder and decoder to create clearer and more accurate reconstructions.

\section{Method}
\subsection{Overview}
The proposed ODDObjects framework consists of three major components: a data processing pipeline, an autoencoder model, and an anomaly detection algorithm. The data pipeline is designed to configure COCO-style object datasets into a suitable and optimized format for autoencoding models. The model can be any type of autoencoder; in this paper, we consider the standard autoencoder, two convolutional variants, and a memory-augmented convolutional variant. The anomaly detection algorithm is based on the reconstruction error of the autoencoder model and is described in detail in section 3.4.

\subsection{Model Architectures}
For consistency, each layer in all models is followed by a ReLU activation \cite{nair2010rectified}, except for the last layers of the encoders and decoders. All encoders shrink the input to a 32-dimensional latent representation. Let Conv2D($n, k, s$) represent a 2D convolution layer and Deconv2D($n, k, s$) represent a 2D deconvolution layer with $n$ filters, a kernel size of $k$, and a stride size of $s$.

The vanilla autoencoder consists of two fully-connected layers, where one is the encoder and one is the decoder. The convolutional autoencoder consists of an encoder with four convolution layers: Conv2D(32, 3, 2), Conv2D(64, 3, 2), Conv2D(128, 3, 2), Conv2D(256, 4, 2), and a decoder with four deconvolution layers: Deconv2D(256, 4, 2), Deconv2D(128, 3, 2), Deconv2D(64, 3, 2), Deconv2D(32, 3, 2), Deconv2D(3, 3, 1). 

The variational autoencoder and convolutional variational autoencoder have the same respective encoder-decoder architectures, with the same ReLU activations. The last layer of the encoder, however, consists of two 32-dimensional vectors that define the latent state distribution, where one is the mean and one is the variance. Using the reparameterization trick \cite{kingma2015variational}, a sampled 32-dimensional latent representation is passed to the decoder. 

The memory-augmented convolutional autoencoder, shown in figure 1, consists of an encoder with three convolution layers: Conv2D(16, 1, 2), Conv2D(32, 3, 2), Conv2D(64, 3, 2), and a decoder with three deconvolution layers: Deconv2D(64, 3, 2), Deconv2D(32, 3, 2), Deconv2D(16, 1, 2). Each layer, other than the last layers of the encoders and decoders, is followed by batch normalization \cite{ioffe2015batch} and a ReLU activation, as before. The memory module is sandwiched between the encoder and decoder, as previously described, and has a memory size of 500.

\subsection{Dataset setup and preprocessing}
COCO 2017 \cite{coco} is a large-scale object detection dataset with 118,000 training images and 5,000 validation images. The images in the dataset contain common objects in their natural surroundings, with 80 object categories. Due to resource constraints, we use an 85/15 split to partition the validation set into 4250 training images and 750 validation images.

\begin{figure}[ht]
\vskip 0.1in
\begin{center}
\centerline{\includegraphics[width=\columnwidth]{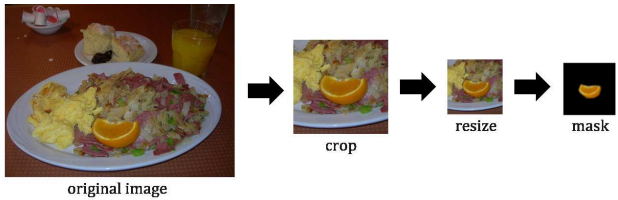}}
\caption{Data preprocessing pipeline.}
\label{data_preprocess}
\end{center}
\vskip -0.25in
\end{figure}

Objects in the dataset are extracted by centering, cropping, and masking based on the location and shape of the object segmentation. It is important to note that a single image may contain multiple objects. The resultant image is a 256x256x3 colour image with the object in the center and the background removed. The background context of the object was removed to focus the models on reconstructing solely the objects without the need to learn object contexts. This significantly improves reconstruction results as well as model performance and convergence. 

\begin{figure}[ht]
\vskip 0.05in
\begin{center}
\centerline{\includegraphics[width=\columnwidth]{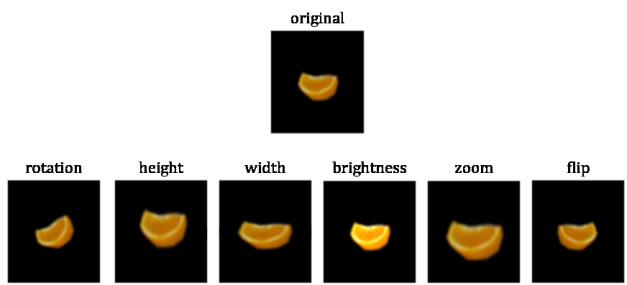}}
\caption{Data augmentations.}
\label{data_aug}
\end{center}
\vskip -0.25in
\end{figure}

The training dataset is augmented by applying a series of random transformations. Each training image is susceptible to a random rotation within 5 degrees, width and height shifts within 1\% of the original width and height, brightness shift within 10\% of the original brightness, zoom within 100\%-125\% of the original image, and a horizontal flip.

\subsection{Memory-augmented autoencoders for anomaly detection}
Given an input $\mathbf{x}$ at inference time, the encoder converts $\mathbf{x}$ to its encoded representation $\mathbf{z}$. The memory module in the memory-augmented autoencoder reconstructs the input using a restricted number of representations recorded in memory, such that $\hat{\mathbf{z}}$ is obtained from the retrieved representations. $\hat{\mathbf{z}}$ is passed to the decoder to obtain a reconstruction of the input, $\hat{\mathbf{x}}$. The reconstruction error equation (10) results in normal samples having small reconstruction errors, while anomalies have large errors. The resulting errors are normalized to have a probabilistic interpretation, and a good threshold was found empirically to be around $0.9$. \textbf{Algorithm 4} shows the anomaly detection algorithm using the MemCAE reconstruction error.

\begin{algorithm}[tb]
   \caption{Anomaly detection with MemCAE}
   \label{alg:anomaly_detection}
\begin{algorithmic}
   \STATE {\bfseries Input:} Dataset $\{X^i\}_{i=1}^N$, threshold $\gamma$, hyperparameter $\alpha$
   \STATE Train autoencoder until parameters $\theta_e$ and $\theta_d$ converge
   \FOR{$i=1$ {\bfseries to} $n$}
   \STATE $\hat{x}^i$ = reconstruction from autoencoder
   \STATE $\textit{error} = ||x^i - \hat{x}^i||_2^2 + \alpha E(\hat{w}^i))$
   \IF{$\textit{error} > \gamma$}
   \STATE $x^i$ is an anomaly
   \ELSE
   \STATE $x^i$ is not an anomaly
   \ENDIF
   \ENDFOR
\end{algorithmic}
\end{algorithm}

\section{Experimental Results}
\subsection{Overview}
In this section, we validate the proposed framework for anomaly detection on segmented objects. To show the generality and applicability of the method, experiments are conducted on the COCO dataset using both single-class and multi-class datasets. We train and compare the results of autoencoders (AE), convolutional autoencoders (CAE), convolutional variational autoencoders (CVAE), and memory-augmented convolutional autoencoders (MemCAE). We use the model architectures described in the previous section, which are implemented using the Tensorflow framework \cite{tensorflow} and trained using the Adam optimizer \cite{kingma2017adam} with a learning rate of 0.001 on an NVIDIA RTX 2080Ti graphics card.

\subsection{Training}
Training on the \textit{orange} class was run for 5,000 steps. Figure 4 shows the training curves for each model, plotting the validation reconstruction loss. The vanilla autoencoder model converged to the highest reconstruction loss, meaning it performed the worst at reconstructing the input images. The convolutional autoencoder and convolutional variational autoencoder performed similarly. The memory-augmented convolutional autoencoder performed the best, converging to a very low reconstruction loss.

\begin{figure}[ht]
\vskip 0.2in
\begin{center}
\centerline{\includegraphics[width=\columnwidth]{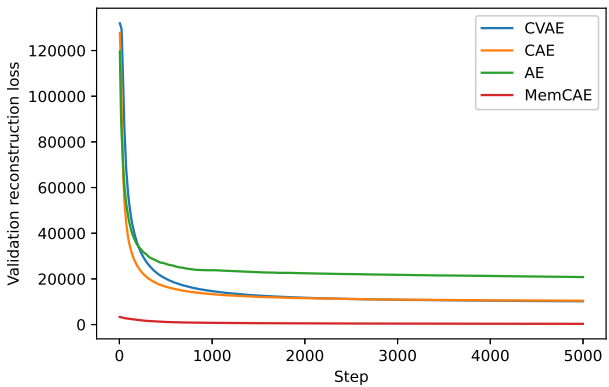}}
\caption{Training curves tracking the reconstruction loss on the validation set with models trained on the \textbf{orange} class.}
\label{training_orange}
\end{center}
\vskip -0.2in
\end{figure}

\begin{figure}[ht]
\begin{center}
\centerline{\includegraphics[width=\columnwidth]{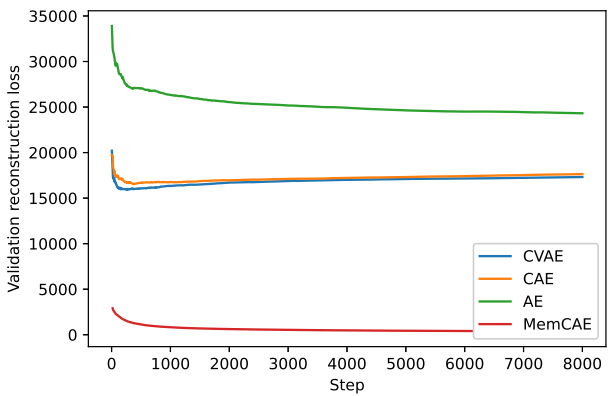}}
\caption{Training curves tracking the reconstruction loss on the validation set with models trained on the \textbf{orange, banana, apple, broccoli, and carrot} classes.}
\label{training_multi}
\end{center}
\vskip -0.2in
\end{figure}

Training on the multi-class dataset was run for 8,000 training steps. The multi-class dataset consists of the \textit{orange}, \textit{banana}, \textit{apple}, \textit{broccoli}, and \textit{carrot} classes. From figure 5, we can see again that the 2-layer autoencoder performed the worst. Likewise, the convolutional autoencoder and convolutional variational autoencoder performed similarly, with both models starting to overfit around step 500. Again, the memory-augmented convolutional autoencoder significantly outperformed all the other models.

\pagebreak
\subsection{Performance evaluation}
\begin{table}[t]
\caption{Experimental results on image data. Average AUC values from models trained on single-class anomaly detection dataset sampled from COCO are shown.}
\label{sample-table}
\vskip 0.15in
\begin{center}
\begin{small}
\begin{sc}
\begin{tabular}{lcccr}
\toprule
Model & AUC\\
\midrule
AE        & 57.49\\
CAE       & 58.89\\
CVAE      & 57.18\\
MemCAE    & \textbf{60.94}\\
\bottomrule
\end{tabular}
\end{sc}
\end{small}
\end{center}
\vskip -0.1in
\end{table}

For the single-class dataset, an anomaly detection dataset is constructed by sampling objects from one class as normal samples and sampling anomalies from the other classes. In the case of the COCO dataset, \textit{orange} was treated as the normal class, while the other 79 classes were treated as anomalies. A total of 5000 objects were sampled to create the dataset. Labeling anomalies as positives and normal samples as negatives, the Area Under Curve (AUC) can be used as a measurement for performance evaluation. This is calculated by finding the area under the Receiver Operation Characteristic (ROC). Table 1 shows a comparison of the different models and their respective average AUC scores. For each model, a series of hyperparameter thresholds for anomaly detection were tested, and the threshold resulting in the best average AUC was chosen.

Although the autoencoder model is not able to reconstruct the input data as adequately as the other models, it performs surprisingly well, achieving an AUC score higher than the convolutional variational autoencoder. The memory-augmented convolutional autoencoder achieved the highest AUC score. While it performed better than the other three models, the improvement is not as significant as was shown from the training curves in figures 4 and 5. 

\subsection{Experiments on image data}
Figures 8-15 show batches of reconstructions from the validation set with the various models trained on the single-class and multi-class datasets. We can see that the 2-layer autoencoder model performs adequately with small circular objects. However, the reconstructions for larger, uncentered, objects are quite blurry. The convolutional models are better at reconstructing these objects. Reconstructions made by the memory-augmented convolutional autoencoder, however, are much clearer than the results from the other models. From figure 15, we can also see that the multi-class MemCAE model is able to match the colours of the objects and capture many of the finer details - something the other three models were not able to do.

\newpage
\begin{figure}[ht]
\vskip 0.2in
\begin{center}
\centerline{\includegraphics[width=\columnwidth]{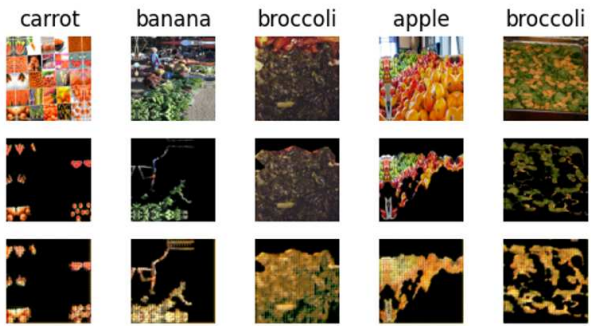}}
\caption{Top 5 most anomalous objects detected by memory-augmented convolutional autoencoder, trained for 10000 steps.}
\label{anomalies_multiclass_memcae}
\end{center}
\vskip -0.2in
\end{figure}

Figure 6 shows the five objects with the highest reconstruction errors, given by the MemCAE model. Evidently, these objects were very poorly labelled, and for many of these, multiple objects are grouped into one segmentation area. The anomaly detection algorithm also lets us intrinsically find the most ``normal" objects in the dataset, which are the objects with the lowest reconstruction errors. These results are shown in figure 7. Comparing the five objects with the highest reconstruction errors and the five with the lowest reconstruction errors, it is evident that the model learned to differentiate between normal and abnormal objects.

\section{Conclusions and Future Work}
This paper introduces ODDObjects, a framework for detecting anomalous objects using deep convolutional autoencoders trained on masked segmentations of objects. We show that ODDObjects can simplify the traditionally tedious data-cleaning process by automatically detecting errors in training datasets without manually reviewing object segmentations one by one. Such a framework can also be scaled for all types of computer-vision-motivated manufacturing processes. Being able to detect anomalies in real-time in autonomous industrial systems ensures the safety and quality of various manufactured products without the need for human supervision.

\newpage
\begin{figure}[ht]
\vskip 0.2in
\begin{center}
\centerline{\includegraphics[width=\columnwidth]{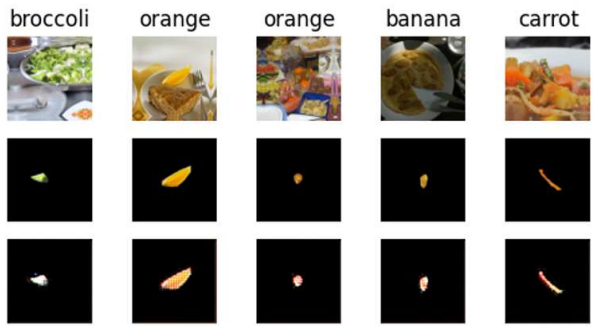}}
\caption{Top 5 most normal objects detected by memory-augmented convolutional autoencoder, trained for 10000 steps.}
\label{normal_multiclass_memcae}
\end{center}
\vskip -0.2in
\end{figure}

\section*{Software}
The code is available at \url{https://github.com/ricky-ma/ODDObjects}. 

\section*{Acknowledgements}
The author would like to thank Peter Hedley for providing supervision, guidance, and patience throughout all stages of the project.

\begin{figure*}[ht]
\vskip 0.2in
\begin{center}
\centerline{\includegraphics[width=\textwidth]{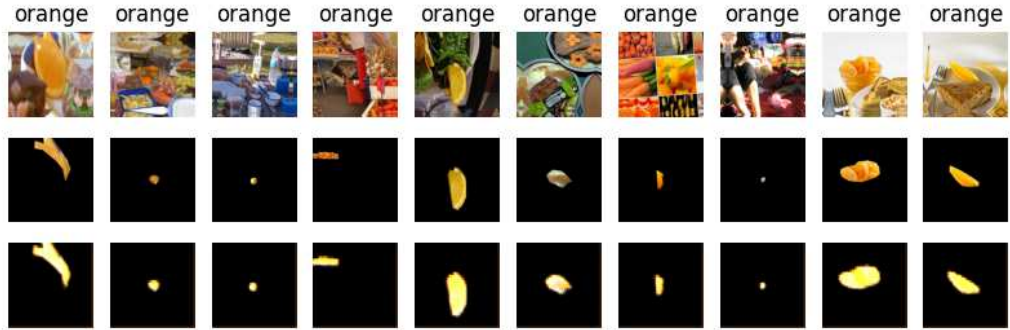}}
\caption{Reconstruction from validation dataset using single-class memory-augmented convolutional autoencoder, trained for 5000 steps.}
\label{memcae_orange}
\end{center}
\vskip -0.2in
\end{figure*}

\begin{figure*}[ht]
\vskip 0.2in
\begin{center}
\centerline{\includegraphics[width=\textwidth]{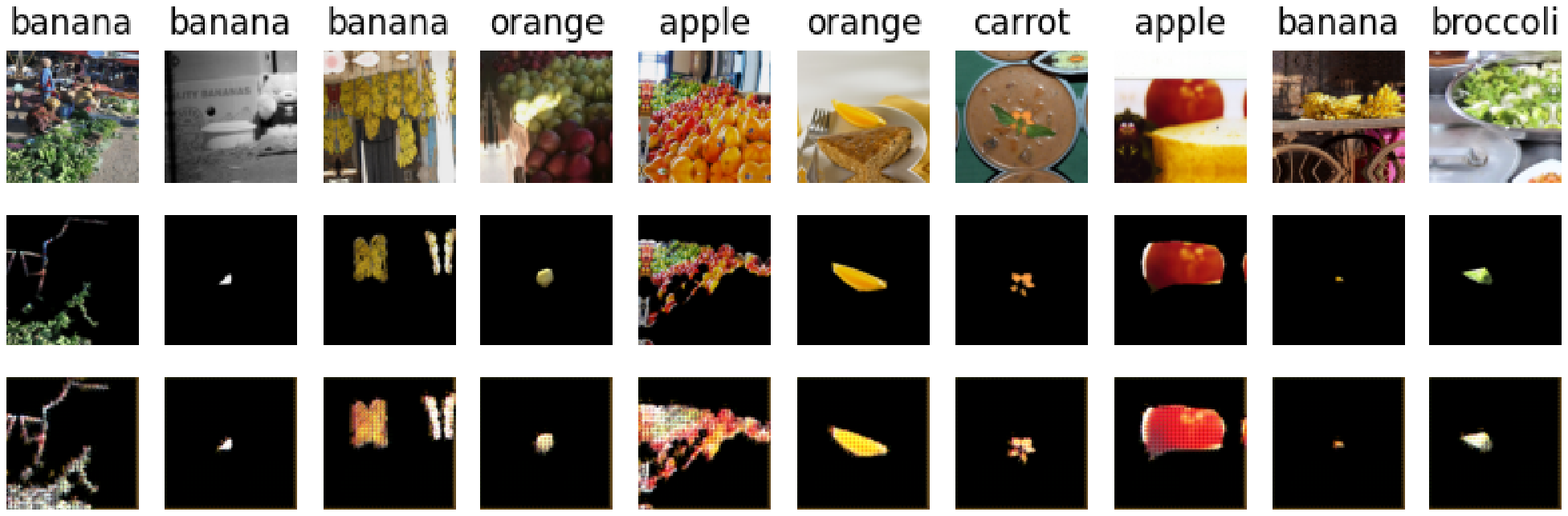}}
\caption{Reconstruction from validation dataset using multiclass memory-augmented convolutional autoencoder, trained for 8000 steps.}
\label{memcae_multi}
\end{center}
\vskip -0.2in
\end{figure*}

\begin{figure*}[ht]
\vskip 0.2in
\begin{center}
\centerline{\includegraphics[width=\textwidth]{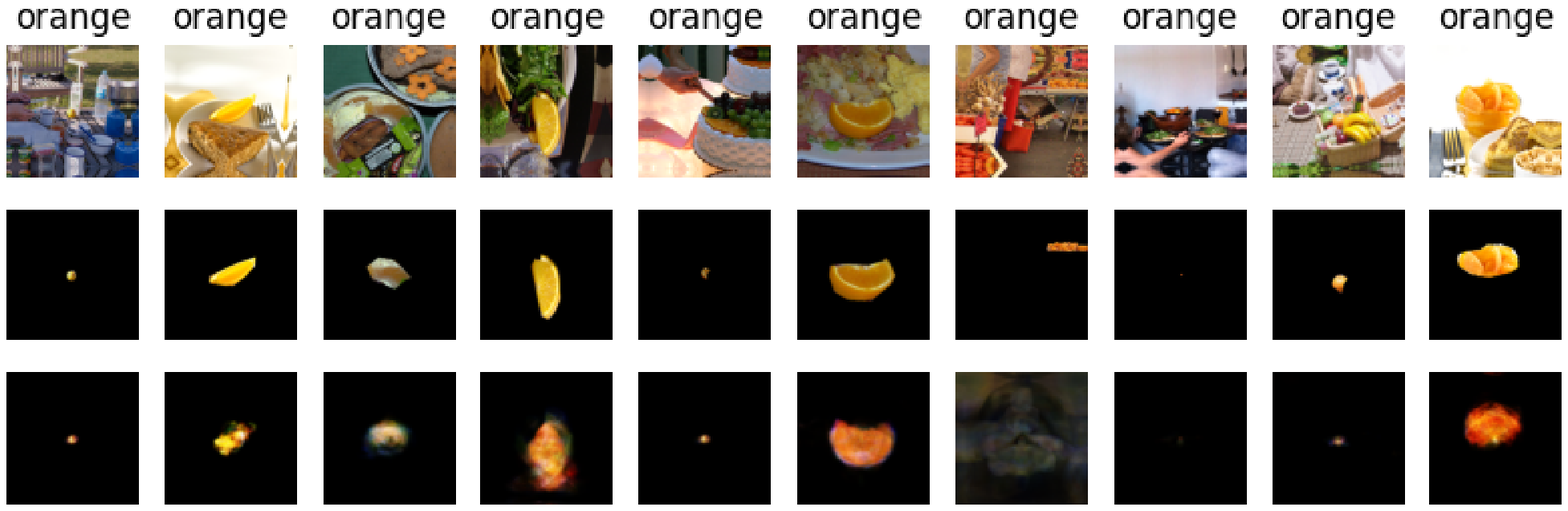}}
\caption{Reconstruction from validation dataset using single-class autoencoder, trained for 5000 steps.}
\label{ae_orange}
\end{center}
\vskip -0.2in
\end{figure*}

\begin{figure*}[ht]
\vskip 0.2in
\begin{center}
\centerline{\includegraphics[width=\textwidth]{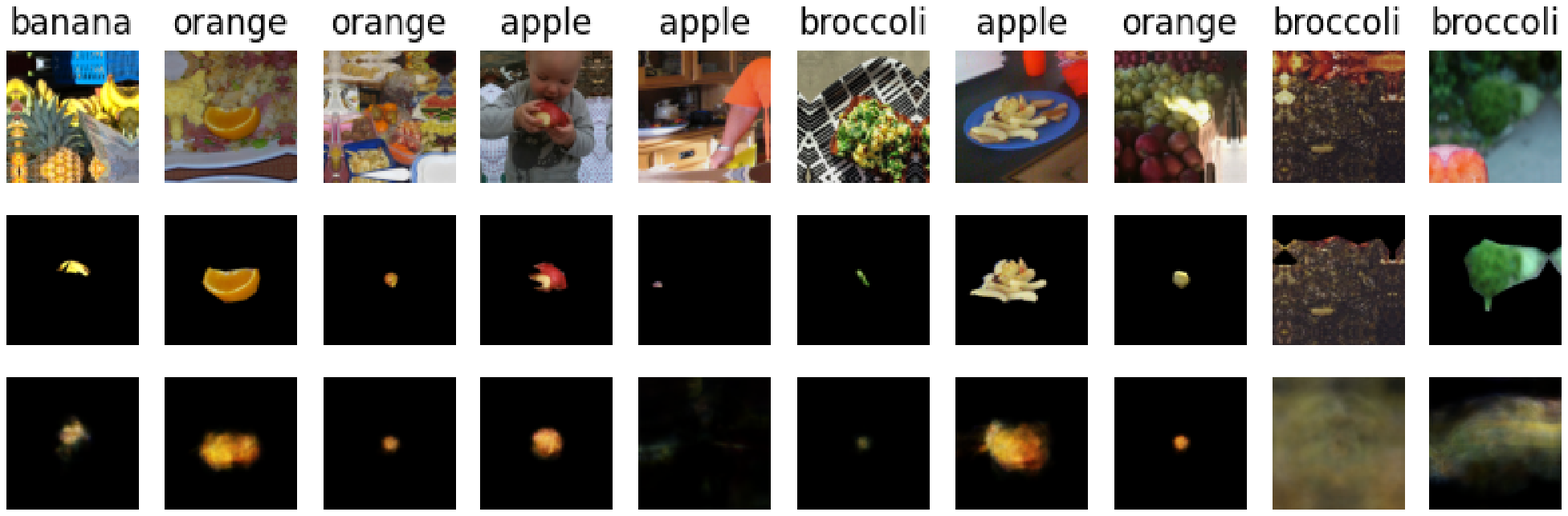}}
\caption{Reconstruction from validation dataset using multi-class autoencoder, trained for 8000 steps.}
\label{ae_similar}
\end{center}
\vskip -0.2in
\end{figure*}

\begin{figure*}[ht]
\vskip 0.2in
\begin{center}
\centerline{\includegraphics[width=\textwidth]{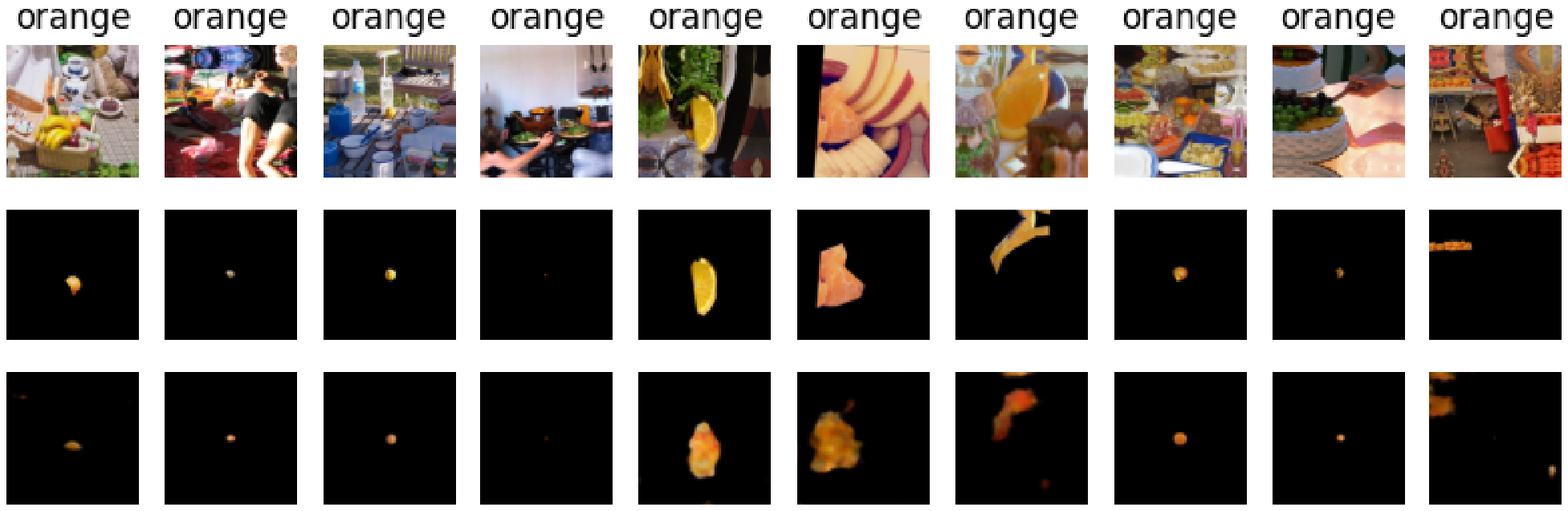}}
\caption{Reconstruction from validation dataset using single-class convolutional autoencoder, trained for 5000 steps.}
\label{cae_orange}
\end{center}
\vskip -0.2in
\end{figure*}

\begin{figure*}[ht]
\vskip 0.2in
\begin{center}
\centerline{\includegraphics[width=\textwidth]{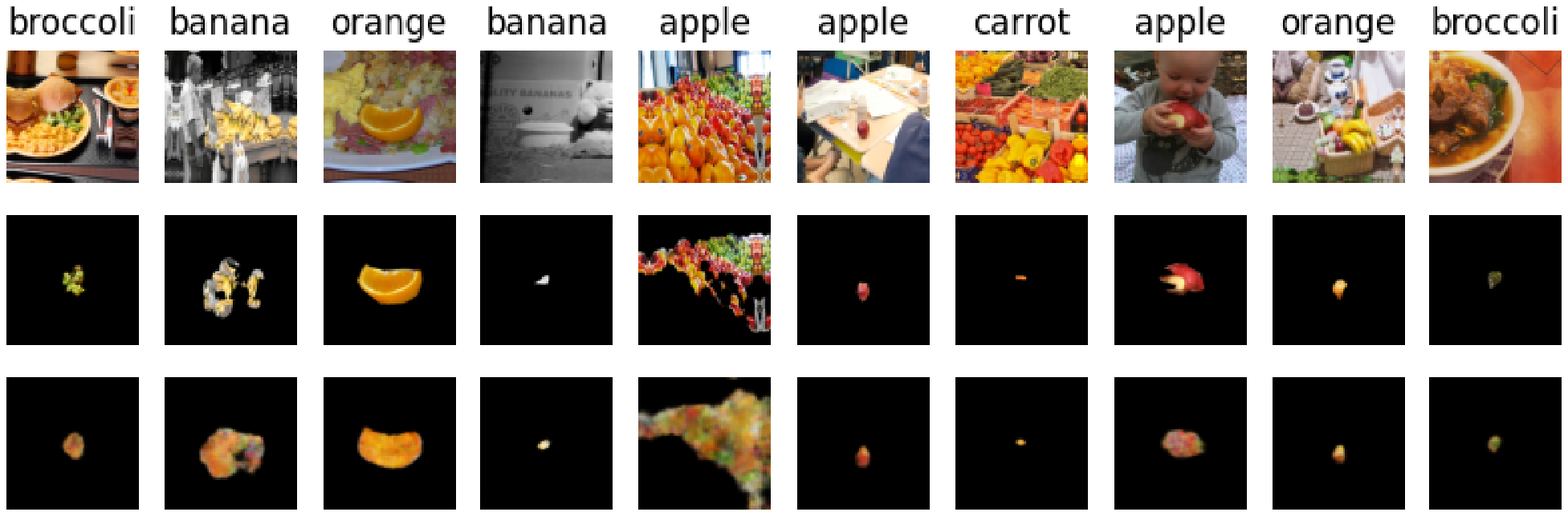}}
\caption{Reconstruction from validation dataset using multi-class convolutional autoencoder, trained for 8000 steps.}
\label{cae_similar}
\end{center}
\vskip -0.2in
\end{figure*}

\begin{figure*}[ht]
\vskip 0.2in
\begin{center}
\centerline{\includegraphics[width=\textwidth]{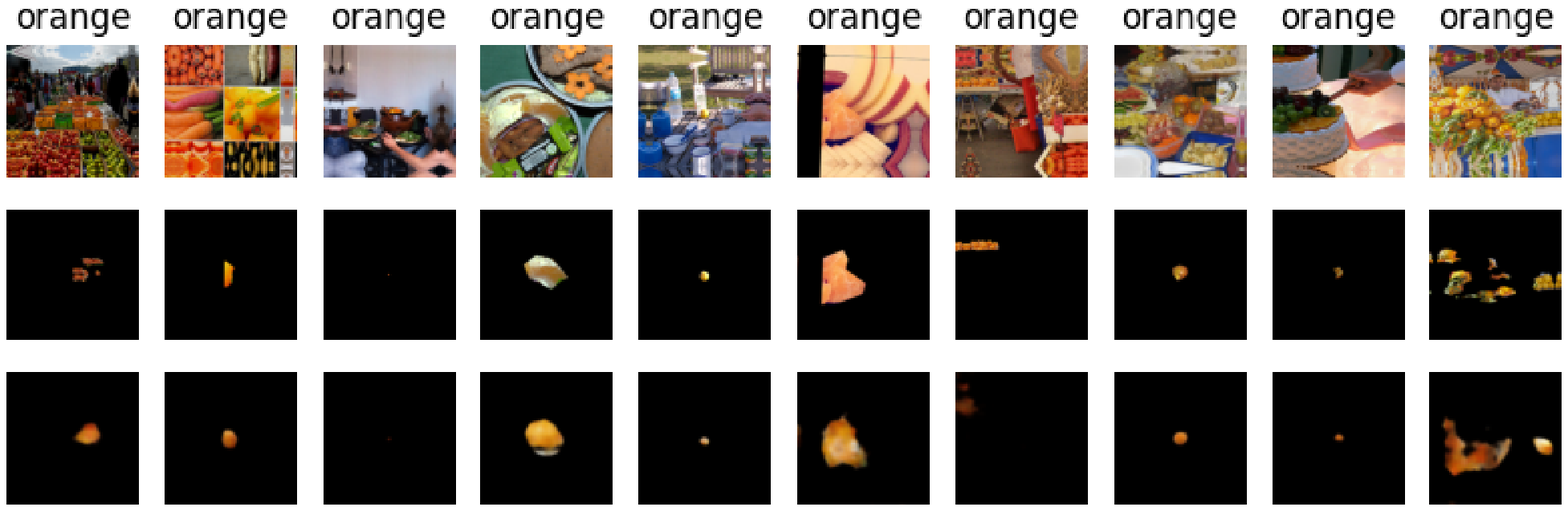}}
\caption{Reconstruction from validation dataset using single-class convolutional variational autoencoder, trained for 5000 steps.}
\label{cvae_orange}
\end{center}
\vskip -0.2in
\end{figure*}

\begin{figure*}[ht]
\vskip 0.2in
\begin{center}
\centerline{\includegraphics[width=\textwidth]{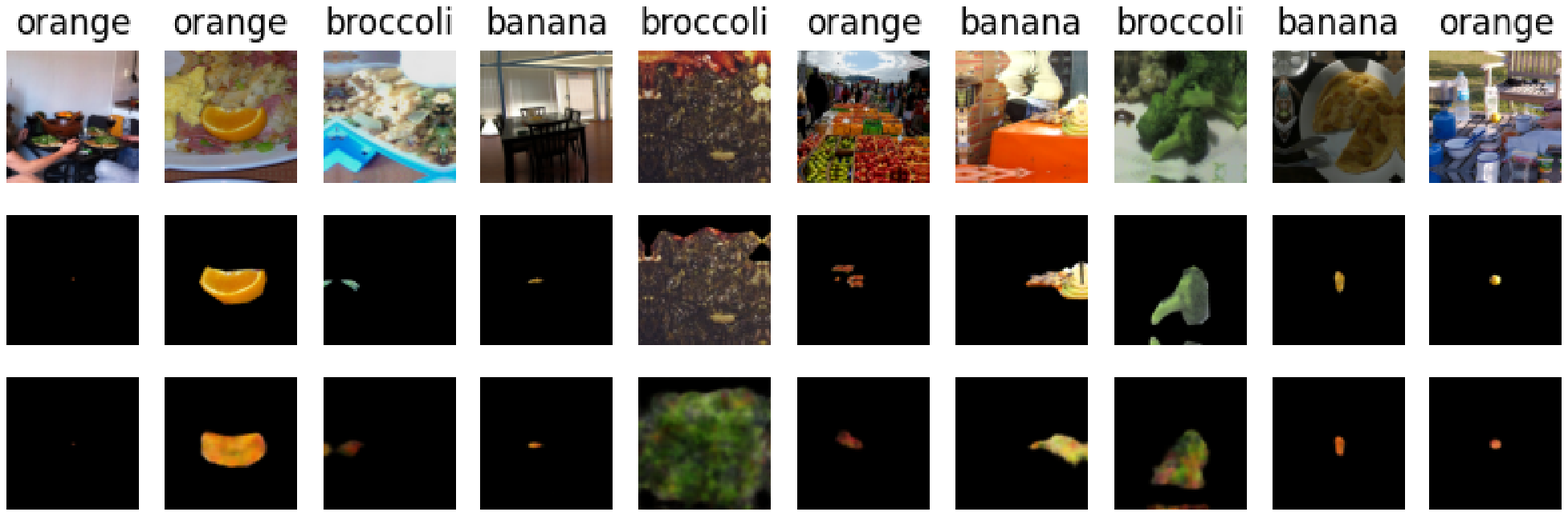}}
\caption{Reconstruction from validation dataset using multi-class convolutional variational autoencoder, trained for 8000 steps.}
\label{cvae_multi}
\end{center}
\vskip -0.2in
\end{figure*}

\clearpage
\clearpage
\bibliography{main.bib}
\bibliographystyle{icml2021}
\end{document}